# DONUT: Physics-aware Machine Learning for Real-time X-ray Nanodiffraction Analysis

## Authors

Aileen Luo[1,2,*], Tao Zhou[3], Ming Du[1], Martin V. Holt[3], Andrej Singer[2,*], Mathew J. Cherukara[1,*]

## Affiliations

[1]Advanced Photon Source, Argonne National Laboratory, Lemont, IL 60439, USA

[2]Department of Materials Science and Engineering, Cornell University, Ithaca, NY 14853, USA

[3]Center for Nanoscale Materials, Argonne National Laboratory, Lemont, IL 60439, USA

*mcherukara@anl.gov, asinger@cornell.edu, aileenluo@anl.gov

## Abstract

Coherent X-ray scattering techniques are critical for investigating the fundamental structural properties of materials at the nanoscale. While advancements have made these experiments more accessible, real-time analysis remains a significant bottleneck, often hindered by artifacts and computational demands. In scanning X-ray nanodiffraction microscopy, which is widely used to spatially resolve structural heterogeneities, this challenge is compounded by the convolution of the divergent beam with the sample's local structure. To address this, we introduce DONUT (Diffraction with Optics for Nanobeam by Unsupervised Training), a physics-aware neural network designed for the rapid and automated analysis of nanobeam diffraction data. By incorporating a differentiable geometric diffraction model directly into its architecture, DONUT learns to predict crystal lattice strain and orientation in real-time. Crucially, this is achieved without reliance on labeled datasets or pre-training, overcoming a fundamental limitation for supervised machine learning in X-ray science. We demonstrate experimentally that DONUT accurately extracts all features within the data over 200 times more efficiently than conventional fitting methods.

## Main Text

### Introduction

Scanning X-ray diffraction microscopy (SXDM) is an advanced imaging technique that spatially resolves structural information by diffraction contrast. A coherent, nano-focused X-ray beam is raster-scanned across an extended sample, and the diffraction patterns collected at each point elucidate the localized strain gradients and crystal lattice orientation. This characterization method is particularly useful for studying thin films with heterogeneous domains, such as ferroelectrics[1,2], photovoltaics[3,4], magnetic oxides[5,6], and Mott insulators[7,8]. Yet, a critical challenge in interpreting SXDM data is the convolution of the beam shape with the local crystal lattice information. The convergence angle of the probe due to nanoscale focusing optics results in simultaneous dependency of the far-field scattering data on the lattice strain and two rigid body rotations of the unit cell with respect to the normal of the Bragg planes. Conventionally, these components are disentangled by fitting to simulated data, a process that is computationally expensive and highly dependent on data pre-processing.

With the widespread use of two-dimensional pixelated area detectors at synchrotron facilities, X-ray data is acquired in the form of images, which lends well to computer vision methods for processing. Supervised machine learning (ML) has achieved great improvements in analysis speed across many scientific disciplines, including Bragg coherent diffractive imaging (BCDI)[9], X-ray ptychography[10], coherent surface scattering[11], and 4D scanning transmission electron microscopy[12]. Nonetheless, supervised models face a fundamental bottleneck: the requirement for large, well-labeled training datasets. Acquiring labeled data for coherent x-ray scattering techniques can be difficult, as it often requires extensive domain expertise to generate simulated data or vastly time-consuming and computationally costly experimental analysis. Furthermore, simulated data may not be an accurate representation of the complexity of experimental data. The reliance on labeled data limits the flexibility and broader applicability of supervised learning models for X-ray science. Unsupervised learning methods, by contrast, do not require labeled data and can automatically discover patterns within datasets. Techniques such as clustering, dimensionality reduction, autoencoders, and generative models have been applied to coherent X-ray scattering experiments such as BCDI[13], ptychography[14], computed tomography[15], and X-ray photon correlation spectroscopy[16], illustrating the effectiveness of unsupervised learning combined with a physical model for a specific task in improving analysis efficiency.

For a physical model to guide a neural network's training, gradients must be calculated not only through the network's parameters but also through the simulation itself. This is made possible by automatic differentiation (AD), a computational technique that efficiently and accurately computes derivatives of complex functions[17]. AD has recently seen widespread use in computational imaging[18], including image reconstruction problems, with workflows specifically developed for coherent X-ray techniques[19,20]. By implementing the forward scattering model in a fully differentiable manner, AD allows the error between the model's prediction and the experimental data to be backpropagated through the entire physics-aware architecture, enabling end-to-end optimization.

In this study, we leverage this approach to develop DONUT: Diffraction with Optics for Nanobeam by Unsupervised Training. DONUT is a physics-aware autoencoder that directly embeds a differentiable SXDM forward model into its architecture. It learns to extract strain and lattice tilt by minimizing the difference between its physics-based diffraction output and the measured experimental data, eliminating the need for pre-generated labels. We demonstrate that DONUT can be trained solely on experimental data, achieving analysis speeds over 200 times faster than conventional methods while providing improved accuracy in disentangling convoluted structural parameters. This framework provides a versatile and robust pathway for real-time, automated analysis of SXDM data, accelerating discovery in the study of complex materials under dynamic conditions.

## Results

**Model approach**

The DONUT architecture is founded on a physics-aware autoencoder. Standard autoencoders are power tools for unsupervised feature extraction[21,22], noise reduction[23], and anomaly detection[24]. These neural networks learn a compact latent space representation of high-dimensional imaging data, capturing essential structural and compositional features while filtering out irrelevant noise. A key challenge with autoencoders, however, is the interpretability of the latent space, as purely data-driven embeddings may lack direct physical significance. To address this, physics-based constraints such as enforced symmetries, conservation laws, or domain-specific priors are incorporated into the network architecture[25,26,27] or loss function[28]. DONUT applies a similar approach by directly building a physics-based X-ray scattering simulation into the forward pass of the neural network and restricting the dimensionality of the latent space to the quantities of interest. By forcing the network to learn through the lens of a physical model, we ensure that the extracted features are not just abstract representations but are directly tied to the structural properties of the material.

As illustrated in Figure 1a, the network takes a 2D diffraction pattern as input. This image is passed through a convolutional neural network (CNN) encoder (Fig. 1b), which compresses it into a low-dimensional latent space vector. By design, this latent vector represents the physical quantities of interest: strain ($\varepsilon$, relative magnitude of the momentum transfer vector $Q$), in-plane rotation ($\omega$, relative rotation angle of $Q$ within the horizontal scattering plane), and out-of-plane rotation ($\chi$, relative rotation angle of $Q$ perpendicular to the scattering plane).

The latent vector then simultaneously feeds into two parallel branches:

1. A **decoder**, symmetric to the encoder, which reconstructs a denoised version of the input diffraction pattern from the latent vector.
2. A **physics-based forward model**, which uses the latent parameters ($\varepsilon$, $\omega$, and $\chi$) to generate a simulated diffraction pattern.

The network is trained by minimizing a composite loss function (See Methods, Neural network architecture and training for details). This loss is the weighted mean absolute error between the original input image and the outputs of both branches (the decoder reconstruction and the physics-based simulation). During training, the network iteratively adjusts its weights until both generated images match the input data. This dual-objective optimization ensures that the encoder learns to populate the laten space with physically accurate and robust values for the desired structural parameters.

The effectiveness of DONUT depends on its physics-based forward scattering model. This model is initialized with a set of fixed experimental and sample parameters, which include:

- **Sample Properties:** The bulk out-of-plane lattice parameter and film thickness, measured prior to the experiment by lab-source X-ray diffraction and X-ray reflectivity, respectively.
- **Diffraction Geometry:** Bragg peak (incident and exit beam angles), X-ray energy, sample-detector distance, and detector pixel size.
- **Focusing Optics:** Zone plate module dimensions and outermost zone width, which define the convergent nature of the incident beam.

These static parameters are used to construct the reciprocal space coordinates for the detector frame. During the forward pass, the model takes the predicted strain and tilt values from the encoder's latent space to define the orientation and shape of the 3D Bragg peak in reciprocal space. The final diffraction pattern is then calculated by projecting the intersection of the 3D Bragg peak with the Ewald sphere construction onto the 2D detector, accounting for the beam's angular divergence (Fig. 1a). by embedding this entire, differentiable physical process into the network, we force the latent space to correspond to concrete physical quantities, ensuring the model's predictions are not just computationally efficient but also scientifically interpretable.

**Performance on simulated data**

To directly compare the performance of DONUT against the conventional correlation analysis method, we simulate features in strain and lattice rotation across a real-space coordinate system indexed $(i,j)$. These spatially distributed ground truth features are then used to generate simulated diffraction patterns representative of an SXDM measurement. Each pixel in the sample grid is associated with one diffraction pattern described by $I_{ijxy}(\varepsilon_{ij}, \chi_{ij}, \omega_{ij}, \boldsymbol{Q}_{xy})$, where $(x,y)$ denote the detector coordinates (see Methods, Physics-informed forward model). Figure 2 shows the comparison between DONUT predictions and the conventional correlation fitting analysis[29]. All the data were simulated to have a maximum intensity of seven photons per pixel (according to the average maximum counts per pixel of the experimental data presented in the following section) before introducing noise by sampling from a Poisson distribution. In the conventional correlation analysis method, the correlation between a simulated diffraction library of 41 values each in strain, in-plane rotation, and out-of-plane rotation, and each of the diffraction patterns in the simulated SXDM scan was calculated by a weighted sum interpolation according to the technique described in Methods, Experimental data analysis. Figure 2b shows that features in strain and in-plane lattice rotation are systematically, incompletely disentangled by conventional analysis, with the periodicity of strain features markedly visible in the in-plane lattice rotation map (middle) and vice versa. Conversely, DONUT (Fig. 2c) cleanly separates the strain and in-plane lattice rotation signal, as well as correctly predicting the magnitude of strain, which is overestimated by conventional analysis. Additional statistics on model prediction accuracy are provided in the Supplementary Information, Figure S1 and Table 1. DONUT is not only more computationally efficient than fitting the correlation between the data and simulation library, but also qualitatively more accurate and quantitatively more precise (SI, Table 1).

**Performance on experimental data**

We next evaluate DONUT on experimental SXDM data from a $SrIrO_3$ thin film and compare its performance against both conventional correlation fitting and a previously developed supervised deep learning model[30]. As shown in Figure 3, DONUT's predictions (Fig. 3b) are qualitatively consistent with the conventional analysis (Fig. 3a) but offer significant improvements. Critically, where the conventional method exhibits clear crosstalk artifacts – with rotational features incorrectly appearing the strain map – DONUT provides a clean, independent separation of the parameters. This disentanglement matches the performance of our prior supervised model; however, DONUT further demonstrates its superior fidelity by resolving the fine striped features between the two larger rotational domains present in the diffraction center-of-mass spatial maps, that the supervised approach could not do (*cf.* Ref. 30, Fig. 4, S5). This indicates that the physics-constrained unsupervised approach not only eliminates the need for labeled data but can also achieve a higher level of accuracy in feature extraction. Uncertainty from this physics-informed model architecture was characterized by Monte Carlo dropout[31] (Fig. S2), which shows that the standard deviation of 30 predictions with 10% of model weights randomly set to zero produces no systematic trends in errors. Although the previously reported supervised deep learning model has an overall smaller model uncertainty (lower standard deviation across predictions by multiple models) than DONUT, the former has systematic prediction errors in strain and in-plane lattice rotation, while the latter has uniform prediction errors, indicating no systematic bias between features.

In this study, we extend SXDM analysis further by adding film thickness prediction to the established parameter axes of strain and lattice rotation. SXDM is sensitive to non-uniformities in film thickness of a crystalline thin film through local variations in total intensity and periodicity of the Laue oscillations (changing fringe positions within the detector frame). Many factors may cause local heterogeneities in film thickness in thin films, including imperfectly controlled growth kinetics during deposition[32], growth modes, and substrate properties[33]. Locally non-uniform film thickness may in turn affect materials properties. Figure 4a shows DONUT's predictions of film thickness for simulated (left) and experimentally measured (right) samples. The top left panel is a map of the ground truth thickness values used to simulate diffraction patterns corresponding to each pixel coordinate. The bottom left panel is the corresponding map of predictions by DONUT. The top right panel shows the conventional analysis by fitting the correlation between measured and simulated diffraction of film thickness for the same sample and scan in Figure 3, while the bottom right panel shows DONUT predictions of the $SrIrO_3$ film thickness. The extension of thickness prediction to DONUT's existing prediction axes of strain, in-plane tilt, and out-of-plane tilt makes no changes to the CNN-based autoencoder structure other than to increase the dimensionality of the bottleneck layer to include a fourth parameter. We omit the other parameters here for brevity, but the full predictions may be found in the Supplementary Information (Fig. S3, S4). We introduce minor changes to the scaling of the encoder output nodes (inputs to the physics-based diffraction model) to achieve numerical stability for training. Rather than the previous direct scaling by dividing by set factors, we take the exponent of the natural logarithm of the factors.

Figure S5 shows predictions of thickness on simulated and experimental data by DONUT trained on a combined simulated and experimental dataset. The discrepancy in intensity normalization in the physics-informed forward model with experimentally measured counts results in simultaneous overprediction of thickness on simulated diffraction and underprediction on experimental data because thickness greatly affects the relative total intensity of the forward model output diffraction pattern. Thus, the bottom left and right panels of Figure 4a show predictions made by DONUT trained only on simulated or experimental data, respectively. The noise in the predictions on simulated diffraction, as well as the large error bars on the leftmost parity plot of Figure S3b, demonstrate the relatively high prediction uncertainty at this thickness regime governed by the detector pixel size in reciprocal space (Methods, Physics-informed forward model). In addition to relative total intensity, film thickness also changes the periodicity of the Laue oscillations present in diffraction from a truncated crystal. Both factors affect DONUT training and inference: at low film thickness, the model only has intensity to compare between input and output diffraction; however, at higher film thickness, fringe spacing becomes an additional learned feature. We hypothesize that thickness prediction is therefore more useful for thicker films, where the dips in intensity in the thickness fringes are within the measured detector frame, as demonstrated in Figure S6.

Unlike with strain and lattice rotation, the conventional analysis of film thickness differs greatly from the neural network inference (Fig. 4a, right), with each method offering its own advantages. The correlation fitting result shows a highly uniform film thickness, which is expected of a highly crystalline film synthesized by molecular beam epitaxy; however, the thickness is far overestimated at 136 Angstroms from the 30 unit cells or 120 Angstroms as measured by in situ reflection high-energy electron diffraction (RHEED), a highly accurate method of surface characterization[34], during synthesis. Conversely, DONUT more reasonably predicts the film thickness to be around an average of 110 Angstroms, within the uncertainty window both of diffraction resolution and crystal growth on different regions of the substrate. Nonetheless, the neural network predictions show a diagonal stripe of higher thickness located at the lower portion of the stripe of large lattice rotations. This is consistent with regions of higher total intensity[35], which may indicate higher film thickness, but we cannot completely rule out the possibility of cross-talk between parameter predictions without further experimental measurements outside the scope of this study. For instance, intensity is not the sole proxy for film thickness as the structure factor also depends on distortions and crystal defects. The conventional fitting result and DONUT prediction are only moderately correlated with a Pearson correlation coefficient of ~0.55 (unlike the other parameters, where both analyses are highly correlated), so we recommend testing thickness analysis on films with a higher degree of non-uniformity.

DONUT may be trained on simulated diffraction, experimental data, or a combination of both. The key advantage of this self-supervised learning approach is that it eliminates the need for pre-generating a simulated dataset with labels for training. Practically, this means that DONUT enables not only real-time analysis of crystalline lattice structure from SXDM experiments, but also simultaneous validation of the accuracy of prediction results through the mean absolute error loss between the input experimental and output model simulated diffraction. We demonstrate that although the model may be trained on part of a single measured scan (Fig. S7a), the best results, according to a balance between high accuracy, low prediction uncertainty, and full dynamic range of predicted values, come from combining noisy simulated and experimental data during the training process (Fig. S7b). We hypothesize that augmenting the experimental data with simulations is likely effective for our specific experimental case because the measured diffraction has a limited range of features caused by strain and lattice tilting. The physics model within DONUT thus only learns the distribution across features present in the input training experimental data and cannot extrapolate to potential outliers during inference. By adding simulated data to the model training, we can ensure learning of a sufficient range of features to cover the solution space of all experimental measurements on a sample, which produces more accurate predictions[36,37].

Nonetheless, training convergence may be reached up to six times faster using the smaller experimental training dataset instead of a combined dataset (see Methods, Neural network architecture and training). Training on noisy simulated data also requires more hyperparameter tuning and does not converge as easily compared to using experimental data; however, this is mitigated by using a combined dataset of clean simulated diffraction and experimental data (adds noise), which also produces accurate predictions with slightly higher uncertainty (Fig. S7b) while still reaching convergence quickly (Fig. S8). Therefore, a realistic workflow at the instrument may involve training the model on the first 20,000 diffraction patterns collected under experimental conditions, deploying the trained model on an edge device to perform inference on the following measurements, continually training the model as more data is collected using other computing resources, and periodically updating the local encoder for real-time predictions, such as illustrated schematically in Figure 4b and previously demonstrated for X-ray ptychography[38]. Final structural analysis may then be refined post-experiment by further augmenting the training dataset with simulated diffraction if necessary.

## Discussion

While AD offers crucial flexibility of model architecture design and integration of physics awareness, scaling of the encoder output nodes that make up forward model inputs to their physically appropriate orders of magnitude may introduce numerical instabilities in the gradients. Although CNN-based autoencoders are commonly implemented for image compression applications, the fundamental goal of SXDM analysis is not compression but rather extraction of lattice quantities from diffraction data. This requires the physics model to constrain the latent space representation produced by the encoder but theoretically should not require a decoder. Nonetheless, we find empirically that the inclusion of a decoder symmetric to the encoder regularizes the optimization process and enables convergence during training without adding significant computation time per epoch. We hypothesize that a combination of gradient instabilities and poor conditioning caused by the complexity of the physics-based component of the model, as well as the constrained latent space, lead to optimization difficulties. The scaling of encoder outputs to physically relevant ranges directly affects the gradient calculation during the backpropagation step as multiplication of large derivatives, which spike the gradients. Furthermore, despite the convex loss landscape (Fig. S9), our choice in activation function for the encoder prediction head also stabilizes gradients. A modified hyperbolic tangent[39] softly constrains the encoder output, which prevents arbitrary predictions while simultaneously discouraging the vanishing gradients of more restrictive activation functions such as the symmetric sigmoid hyperbolic tangent, piecewise hard tanh, and standard logistic functions. Thus, the decoder likely plays a similar role in regularizing the learned features by the encoder, as suggested by the evident denoising of the input image by the full autoencoder (Fig. 1a). By disproportionately heavily weighing the physics component of the loss function (See Methods, Model architecture and training) against the decoder component, the encoder output is still majority governed by the forward scattering model. Additionally, extreme sensitivity to the learning rate during training supports the case for adding a decoder (trained at a lower learning rate than the encoder), as the range of viable learning rates is very narrow even with a decoder and the authors were unable to find a reasonable solution without.

DONUT significantly improves the computational efficiency of analyzing SXDM data compared to conventional methods. Using a single NVIDIA GeForce RTX 3090 GPU, the conventional correlation fitting analysis requires $5.6 \pm 0.4$ ms per frame, where each frame is 64 x 64 pixels in single-precision floating-point format (32 bits). The computation time scales linearly with the number of data points and the bottleneck is memory, not including the time required to generate a simulated diffraction library. Inference by the DONUT encoder, conversely, takes $0.024 \pm 0.001$ ms, over 230 times faster than conventional analysis using the same GPU hardware and comparable to the $0.019 \pm 0.001$ ms previously reported for NanobeamNN, the supervised deep learning approach. On CPU, the DONUT encoder predicts structural parameters from diffraction data at a rate of $0.27 \pm 0.07$ ms per frame, more than fast enough to keep up with the maximum detector acquisition rate of 1 kHz without overhead. Thus, the proposed experimental analysis workflow outlined in Figure 4b allows for local inference on CPU by the encoder and simultaneous training of the full model on GPU, constantly updating the model with the most recent data while offloading costly backpropagation to separate hardware and preventing data transfer bottlenecks. Although NanobeamNN has 20% faster inference and roughly 75% faster training than DONUT, NanobeamNN requires pre-simulating a diffraction library, whereas DONUT may be trained on experimental data alone. This key advantage of DONUT, which trains in a self-supervised manner, allows for more flexible extension to additional prediction parameters, such as in the case of the film thickness shown in Figure 4, because the model may be retrained without generating additional simulated data. Adding parameter axes to simulated data generation increases the computational time for that step exponentially, as well as increasing the model retraining time due to training dataset expansion, while DONUT may be retrained on the same experimental dataset. Therefore, DONUT offers a greater degree of data analysis customization during an SXDM experiment by combining physics-based analytical methods with deep learning accelerated optimization compared to both conventional methods and supervised learning.

In conclusion, this work introduces DONUT, a physics-aware unsupervised deep learning framework designed for the rapid and accurate analysis of scanning X-ray nanodiffraction microscopy data. By integrating a differentiable geometric diffraction model directly into an autoencoder architecture, DONUT successfully extracts crystal lattice strain and orientation information in real time, achieving analysis speeds significantly faster than conventional correlation fitting methods without the need for labeled training data. We demonstrate that DONUT not only matches the accuracy of traditional methodologies but also offers improved disentanglement of convoluted structural features and can be effectively trained on experimental data with customization of prediction axes or augmented with simulations for enhanced dynamic range and robustness. This approach significantly lowers the barrier for real-time feedback during complex nanodiffraction experiments, paving the way for accelerated understanding of fundamental materials properties and enabling the study of nanoscale dynamic processes across various scientific domains.

## Methods

**Neural network architecture and training**

DONUT is a convolutional neural network based autoencoder with a physics-informed forward model that predicts crystalline lattice information from scanning X-ray diffraction microscopy data. The model, implemented in PyTorch, can be trained on simulated diffraction, experimental data, or a combination of both. As shown in Figure 1b, the encoder compresses information from 2D diffraction patterns into a latent space of three components and is comprised of blocks of convolution, batch normalization, rectified linear unit (ReLU), dropout, and maximum pooling layers. These blocks are followed by a fully connected layer and modified hyperbolic tangent activation[39] $f(x) = 1.7159 \tanh(\frac{2}{3}x)$, which imposes a soft constraint on the range of values in the bottleneck layer. The bottleneck layer, or latent space representation, is passed to the decoder, which is symmetric to the encoder and acts as a regularization for the optimizer across the loss landscape. Additionally, the physics-based X-ray nanodiffraction scattering model is incorporated into the forward pass of the neural network and takes the latent space representation as inputs. All elements of the latent space are weighted equally for the autoencoder and scaled appropriately to sensible ranges for expected lattice values for the scattering model. Thus, the full DONUT has three final outputs: 1) the latent space tensor of input diffraction consisting of strain, in-plane lattice rotation, and out-of-plane lattice rotation elements, 2) the decoder-reconstructed image of the input (denoises the input), and 3) the simulated diffraction pattern from the physics-aware diffraction model. The loss function for training is a custom weighted mean absolute error (MAE) between the input diffraction and the latter two model output images. The final target outcome for prediction is the tensor of accurate lattice strain and orientation.

The full loss function is formulated as:

$$\text{WMAE}(x_i) = \frac{1}{n}\sum_{i=1}^{n} w_d |x_i - D_{x_i}[E_\theta(x_i)]| + w_f |x_i - f[E_\theta(x_i)]|;$$

where $x_i$ is the input diffraction patterns, $n$ is the number of images in the dataset, $E_\theta$ is the encoder with parameters $\theta$, $D_{x_i}$ is the decoder, $f$ is the physics-based forward model, $w_d$ is the relative weight of the decoder MAE loss (set to 1 in our implementation), and $w_f$ is the relative weight of the physics model loss (set to 5).

The training dataset of simulated diffraction patterns spans ranges of 41 images each in strain $\varepsilon$, in-plane lattice tilt $\omega$, and out-of-plane lattice tilt $\chi$, for a total of 68,921 images of 64 x 64 pixels, where $\varepsilon \in [-0.005, 0.005]$, $\omega \in [-0.05°, 0.05°]$, and $\chi \in [-0.1°, 0.1°]$ (appropriate values for structural heterogeneities often observed in experiments). Each simulated diffraction pattern is scaled to match experimentally measured intensities, then sampled from a Poisson distribution for realistic noise approximation. The experimental dataset consists of one scan of 165 x 165 positions across a sample, where a 128 x 128 pixel region of interest is selected on the detector and downsampled to 64 x 64 pixels to improve the signal-to-noise ratio. The final working dataset (simulated, experimental, or combination) is randomly split into training (80%), validation (10%), and test (10%) sets. The training cycle uses the adaptive moment estimation optimizer with decoupled weight decay regularization[40] (AdamW) with a learning rate of $10^{-4}$ for the global (encoder + forward model) weights and biases and a learning rate of $10^{-5}$ for the decoder parameters. These learning rate values are for DONUT trained on simulated data and require adjustment for training on experimental and combined data. A dropout ratio of 0.1 is applied to both the encoder and decoder to characterize the model error by Monte Carlo dropout (here 10% of the weights are randomly turned off in both training and testing), but dropout is not recommended in any final models for analysis due to slowing down the training process and increasing the difficulty of reaching convergence. We do not observe significant overfitting without dropout (Fig. S7b), likely due to the strong physics-based constraint from the forward scattering model. The network trains on four NVIDIA GeForce RTX 3090 GPUs for 30 epochs, which takes four hours with a batch size of 16. Alternatively, training exclusively on experimental data can be accomplished with the same hardware in 40 minutes.

**Physics-informed forward model**

The neural network is initialized with constants pertaining to the bulk sample information (lattice constant, Bragg reflection, and film thickness) as well as the instrument parameters (photon energy, detector pixel size, sample-detector distance, and zone plate information). Model initialization also includes construction of the detector reciprocal space and zone plate coordinate systems, which are required for the diffraction simulation but remain constant for one single-angle SXDM experiment and are therefore not trainable model weights. For a detailed derivation of the diffraction geometry with a diverging beam caused by zone plate focusing optics, please refer to previous work. Briefly:

Detector coordinates: Each pixel on the 2D area detector is associated with a value of $Q$. Coordinate matrices $X$ and $Y$ are created as 2D meshes, and the detector Q-space mapping is defined below, where the wavevector $k = \frac{2\pi}{\lambda}$, $\theta$ is the angle of the incident X-ray beam, $\gamma$ is the angle of exit beam, $p$ is the detector pixel size, $R$ is the sample-detector distance, and $\tau = k\frac{p}{R}$ is the detector pixel size in reciprocal space (theoretical reciprocal space resolution of the instrument under given conditions).

$$D_{Qx} = k[\cos(\theta) - \cos(\gamma)]$$

$$D_{Qy} = Y\tau$$

$$D_{Qz} = k[\sin(\gamma) + \sin(\theta)]$$

Zone plate coordinates: The angular divergence in the diffracted beam created by the zone plate focusing optics is calculated as a shifting of the origin of reciprocal space (due to varying angles in the convergent incident beam), where the origin $O_Q$ is defined for each coordinate and a binary mask $O_{donut}$ sets the divergence angle boundaries according to the parameters of the optics suite. The complete formulation may be found in previous work, while here we detail a slight numerical optimization to defining the zone plate effects by applying the mask at this step rather than in the final intensity function, reducing the number of dimensions in the calculation. The result, however, remains equivalent across both approaches.

$$Z_{Qx} = O_{Qx}(u,v)[O_{donut}(u,v)] = (o_{Qx,uv})(u,v) \in \{(u,v) | o_{donut,uv} = True\}$$

$$Z_{Qy} = \cdots, Z_{Qz} = \cdots$$

In this work, we introduce optimizations that not only improve upon the computational efficiency of the forward model (and thus greatly cutting down on the backpropagation time[41] using automatic differentiation), but also provide flexibility for customization of sample-specific features. While the coordinate systems defined above are initialized with the neural network, the remainder of the physics-based diffraction model is calculated with the encoder output in every forward pass.

Diffraction intensity function: The Bragg diffraction is defined by the geometry of the crystal along each component of $Q$. For a non-specific specular reflection, the $Q_x$ and $Q_y$ directions correspond to the lattice tilts. Due to the quasi-infinite nature of the plane of a thin film sample, the shape of the Bragg peak along those directions is approximated by very sharp Gaussian functions. The $Q_z$ direction corresponds to the sample norm and the sinc function comes from diffraction by a truncated crystal.

1. Define the components of the momentum transfer vector containing lattice information as meshes of masked projections on the detector (three axes in each component: mask dimension containing the zone plate effects and two detector coordinates). $\varepsilon$, $\omega$, and $\chi$ are the strain, in-plane tilt, and out-of-plane tilt, respectively, scaled from the bottleneck layer of the encoder. $c$ is the bulk out-of-plane lattice parameter in angstroms and $l$ refers to the $00l$ Bragg peak.

$$q_x = D_{Qx} + \frac{2\pi l \omega}{c(1+\varepsilon)} - Z_{Qx}$$

$$q_y = D_{Qy} + \frac{2\pi l \chi}{c(1+\varepsilon)} - Z_{Qy}$$

$$q_z = D_{Qz} - \frac{2\pi l}{c(1+\varepsilon)} - Z_{Qz}$$

2. Define the intensity function as the intersection between three shape functions in $Q$ projected on the detector frame with the angular divergence of the diffracted beam, where $m$ represents the mask axis due to the zone plate effects, and $x$ and $y$ are the detector coordinates.

$$I(\varepsilon, \omega, \chi)_{m,x,y} = \sum_m t \cdot \mathrm{sinc}^2\left(\frac{t}{2\pi} q_z\right) \cdot e^{-q_x^2/\sigma_x^2} \cdot e^{-q_y^2/\sigma_y^2};$$

where:

- $t$ is the film thickness
- $\text{sinc}(x) = \frac{\sin(\pi x)}{\pi x}$
- $\sigma_x$, $\sigma_y$ control the width of the Bragg peak in $Q_x$, $Q_y$, respectively, and thus also serve as a proxy for the crystalline quality of the sample
  - For a highly crystalline, in-plane isotropic thin film:

$$\sigma_x = \sigma_y \approx \frac{\tau}{10}$$

**Automatic differentiation**

Differentiation is essential in machine learning because it provides the mechanism by which models learn: computing gradients of the loss function with respect to model parameters allows optimization algorithms to iteratively adjust those parameters to minimize error and improve predictive performance. Automatic differentiation (AD) computes precise derivatives through computational graphs and the systematic application of the chain rule and is implemented in machine learning libraries including TensorFlow, PyTorch, and JAX. Nonetheless, the smoothness and differentiability of the functions involved significantly affect its effectiveness[42,43]. Smooth, continuously differentiable functions facilitate stable, accurate, and efficient gradient computations, allowing for reliable convergence during optimization. In contrast, using piecewise functions may introduce non-differentiable points or sharp transitions. One of the key differences between the simulated data generation method reported in the supervised deep learning approach to SXDM analysis and the forward scattering model employed by DONUT lies in the differentiability of the 3D Bragg intensity formulation. The previously reported simulation treats the two dimensions of the Bragg peak relating to the extended plane of the thin film as a 2D rectangle function along $\boldsymbol{Q}_x$ and $\boldsymbol{Q}_y$. This comes from the theoretical representation of an infinite crystal in reciprocal space, where the diffraction should be an infinitely sharp Dirac delta function. A real epitaxial thin film is not a perfect infinite crystal, but the lattice is sufficiently ordered at the length scale of the X-ray probe size to approximate the local diffraction as a tightly bound box of intensity. Practically, this means that two piecewise absolute value functions which are not differentiable at the boundaries constrain the Bragg peak in $\boldsymbol{Q}_x$ and $\boldsymbol{Q}_y$ within a sharp rectangle propagated along the $\boldsymbol{Q}_z$ axis into a prism. While this is irrelevant to the supervised approach and does not affect the computational efficiency of the forward pass, backpropagation becomes significantly slower in self-supervised learning where the X-ray scattering model is incorporated into the neural network architecture. Thus, the forward model in DONUT approximates the Bragg peak along $\boldsymbol{Q}_x$ and $\boldsymbol{Q}_y$ as two independent sharp Gaussian functions, which are continuously differentiable and more physics-based approximations of Bragg diffraction from a semi-infinite crystal.

**Data acquisition**

The experimental methods, including sample synthesis and scanning X-ray nanodiffraction microscopy data acquisition for $SrIrO_3$ have been reported previously[35].

**Experimental data analysis**

Experimental data is used to test the performance of DONUT against state-of-the-art conventional data analysis methods. A region of interest (ROI) of detector pixels is selected to encompass the entire measured diffraction peak. The pixels of this ROI are binned by aggregating each 2 x 2 pixel array into one value to improve the signal to noise ratio. A simulated diffraction library of patterns spanning the space of lattice quantities is generated using the physics-based forward scattering model. Each experimentally measured diffraction pattern is multiplied with each simulated diffraction pattern to form a series of correlation matrices. Projections of the correlation matrices are made along each lattice quantity axis and the highest correlation value for each parameter is obtained by a weighted sum interpolation for the center of mass of the projection.

## Data availability

The simulated and experimental datasets that support the findings of this study will be made available in a public repository upon acceptance of this manuscript for publication.

## Code availability

The code and trained model developed in this will be made available in a public GitHub repository upon acceptance of this manuscript and approval from our institution.

## Acknowledgements

This work is supported by the U.S. Department of Energy, Office of Science, Advanced Scientific Computing Research and Basic Energy Sciences award X-ray Scientific Center for Optimization,

Prediction, and Experimentation (XSCOPE), under Contract No. DE-AC02-06CH11357. This work was also supported by the U.S. Department of Energy, Office of Science, Office of Workforce Development for Teachers and Scientists, Office of Science Graduate Student Research (SCGSR) program. The SCGSR program is administered by the Oak Ridge Institute for Science and Education for the DOE under contract number DE-SC0014664. A.S. acknowledges the support by the U.S. Department of Energy, Office of Science, Office of Basic Energy Sciences (Contract No. DE-SC0019414). M.J.C also acknowledge support from the U.S. Department of Energy, Office of Science, Office of Basic Energy Sciences Data, Artificial Intelligence, and Machine Learning at DOE Scientific User Facilities program under Award Number 34532. We gratefully acknowledge the computing resources provided and operated by the Joint Laboratory for System Evaluation (JLSE) at Argonne National Laboratory. Work performed at the Center for Nanoscale Materials and Advanced Photon Source, both U.S. Department of Energy Office of Science User Facilities, was supported by the U.S. DOE, Office of Basic Energy Sciences, under Contract No. DE-AC02-06CH11357.

## Author contributions

A.L. designed the neural network architecture, performed network training and testing, adapted the simulations to experimental conditions, analyzed the data, and expanded the simulations and neural network to additional parameter axes, with advising from M.D., T.Z., A.S., and M.J.C. T.Z. and M.V.H. developed the original simulation and correlation fitting method. A.L., T.Z. A.S., and M.J.C. contributed to the proposal of the initial idea. All authors contributed to the manuscript writing.

## Competing interests

The authors declare no competing interests.

# Figures

**Figure 1: Depiction of the unsupervised deep learning workflow for prediction of lattice strain and tilts from scanning X-ray diffraction microscopy data.**

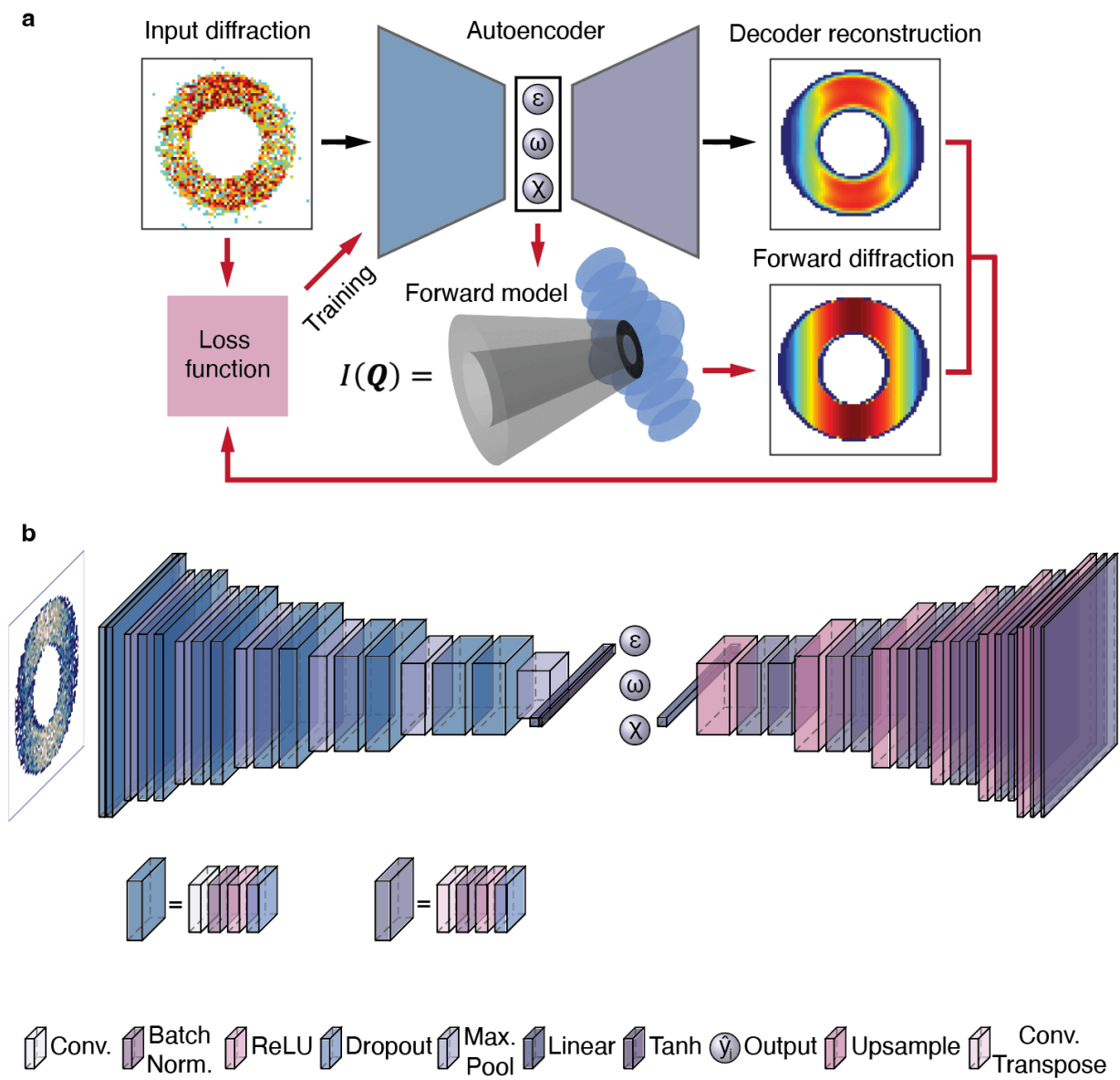

**a** Schematic illustration of the model architecture and self-supervised training process of DONUT. The input diffraction pattern (experimental data shown here) is passed through an autoencoder, and the bottleneck layer, which is the target prediction output, is fed to the physics-based forward scattering model, which is a convolution of an approximation of the convergent probe and the Bragg peak projected onto the detector. The simulated diffraction and decoder reconstruction are compared against the input diffraction in a weighted mean absolute error loss function. **b** Diagram of the layers in the CNN-based autoencoder (box dimensions not drawn to scale w.r.t. layer dimensions).

**Figure 2: Feature extraction from simulated scanning X-ray nanoprobe diffraction microscopy measurement by conventional fitting and DONUT.**

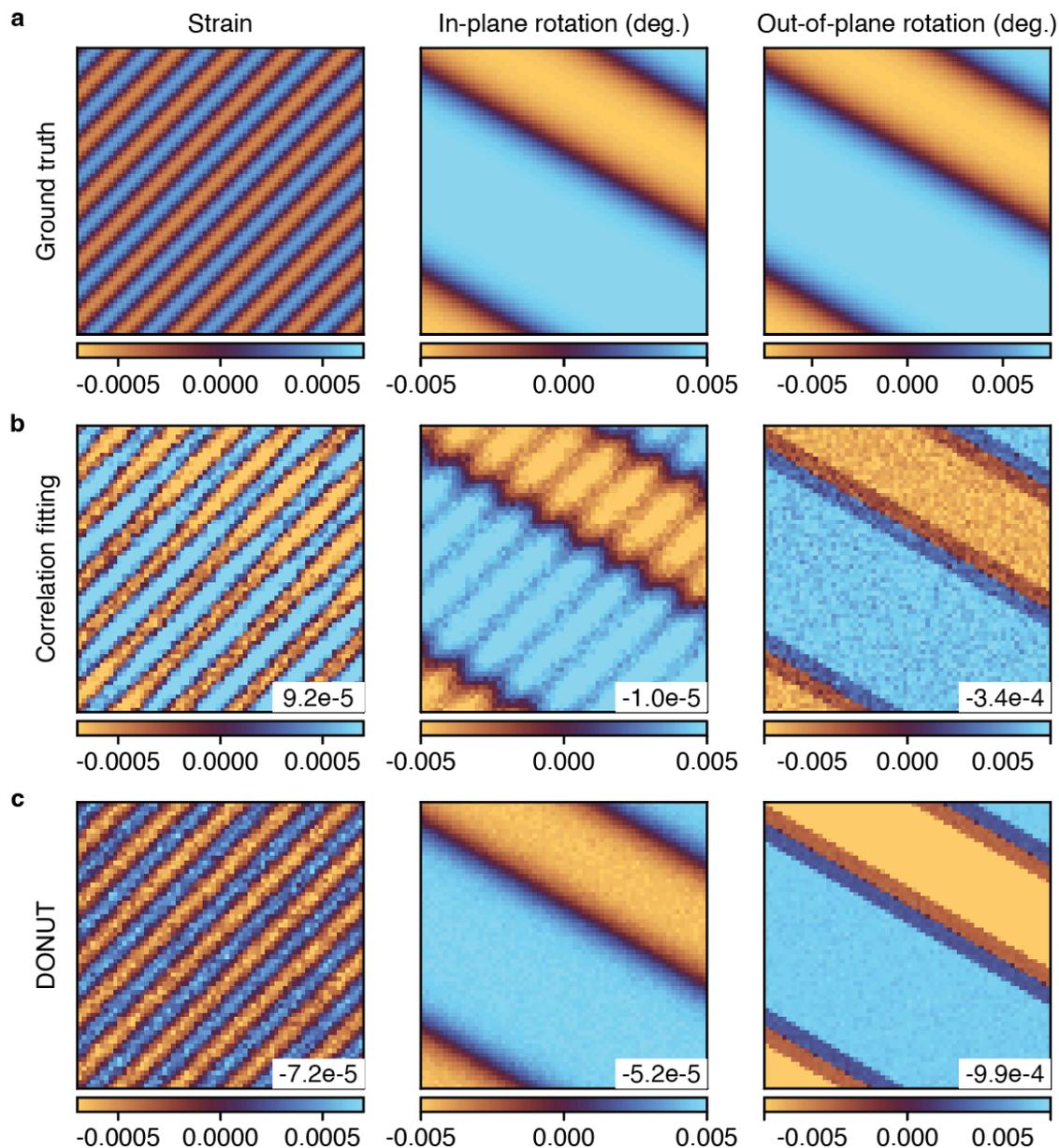

**a** Simulated spatial distribution of strains $\varepsilon(i,j)$ and tilts $\omega(i,j)$, and $\chi(i,j)$, which mimic extended features present in a scanning X-ray nanoprobe diffraction microscopy measurement, where there is one diffraction pattern for each unique pixel coordinate. **b** $\varepsilon(i,j)$, $\omega(i,j)$, and $\chi(i,j)$, as analyzed by conventional fitting of the correlation between measured and simulated diffraction. **c** $\varepsilon(i,j)$, $\omega(i,j)$, and $\chi(i,j)$, predicted by DONUT. The numbers in the bottom right corners of the maps in (**b**) and (**c**) indicate the MAE between the analyzed result and the ground truth.

**Figure 3: Comparison of conventional fitting and DONUT on experimental data.**

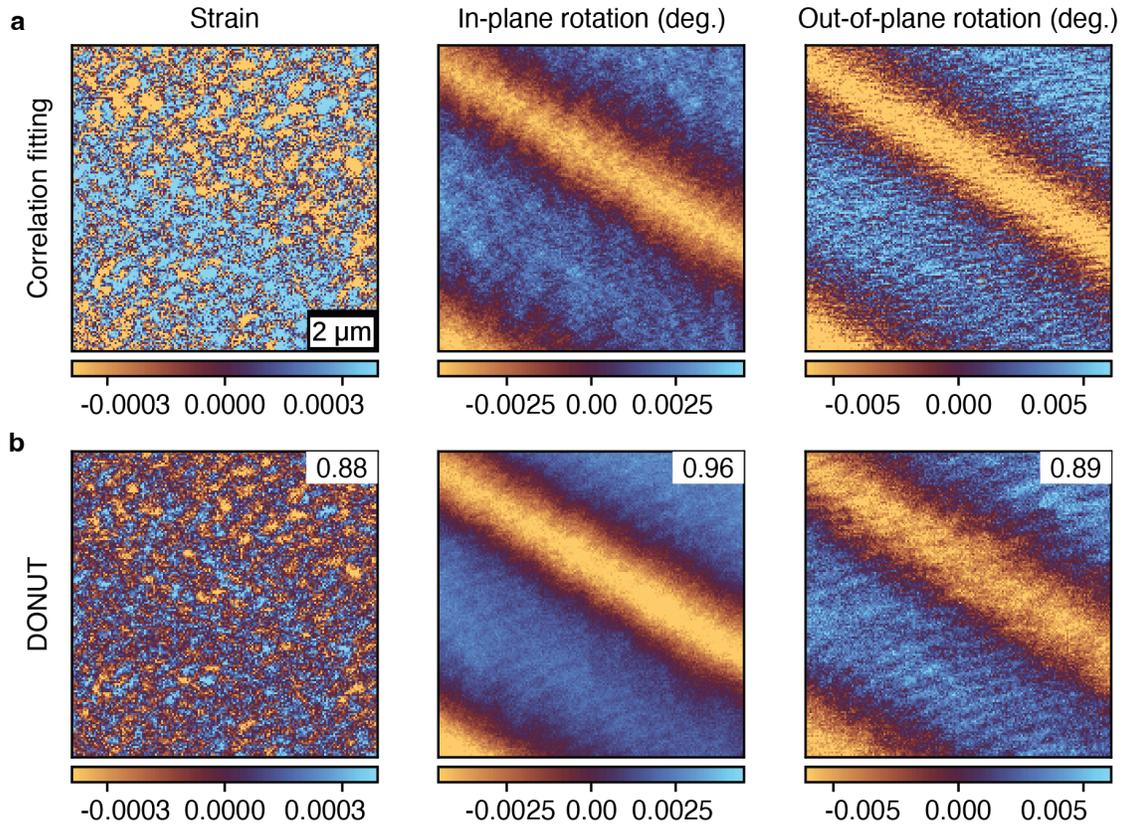

**a** Diffraction contrast features of the SrIrO$_3$ 002$_{pc}$ peak measured by scanning X-ray nanoprobe diffraction microscopy as analyzed by conventional fitting of the correlation between measured and simulated diffraction. **b** DONUT evaluations of the experimental data shown in (**a**). The labels in the top right of the maps are the Pearson correlation coefficients between the two analyses results for each structural parameter.

**Figure 4: Extension of DONUT to predict film thickness and illustration of real-time experimental analysis workflow.**

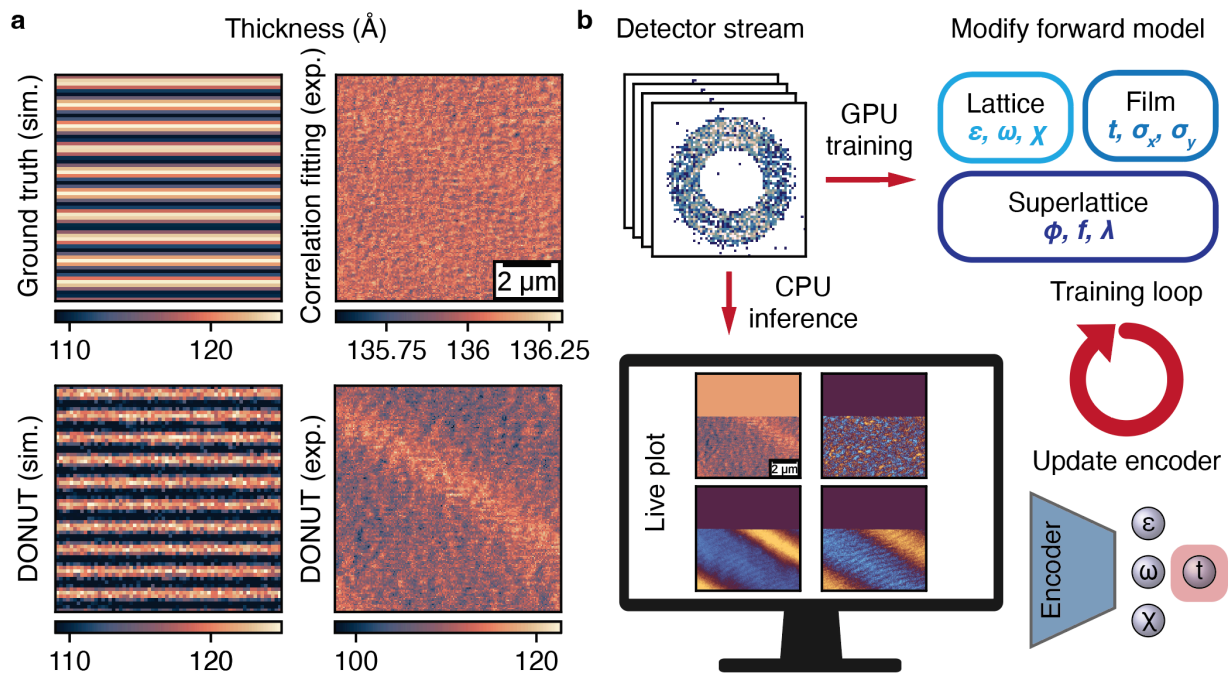

**a** Top left: simulated ground truth values of film thickness $t(i,j)$ across a spatial region of a sample. Bottom left: predicted values of $t(i,j)$ using a version of DONUT trained only on simulated data. Top right: conventional analysis of film thickness from the $SrIrO_3$ $002_{pc}$ peak measured by scanning X-ray nanoprobe diffraction microscopy. Bottom right: predicted values of film thickness by DONUT trained on experimental data. **b** Schematic depiction of real-time analysis workflow during a scanning X-ray nanodiffraction microscopy experiment.